\pdfoutput=1

\documentclass[11pt]{article}

\usepackage{acl}

\usepackage{times}
\usepackage{latexsym}

\usepackage[T1]{fontenc}  
\usepackage[utf8]{inputenc}
\usepackage[english]{babel} 

\usepackage{longtable}
\usepackage{caption}

\usepackage{placeins}

\usepackage{url}
\usepackage{changepage}

\usepackage{hyperref}
\usepackage{url}
\usepackage{graphicx}
\usepackage{multirow}
\usepackage{booktabs}
\usepackage[inline]{enumitem}
\usepackage{caption}
\usepackage{subcaption}
\usepackage{xcolor}
\usepackage{nccmath}

\usepackage{amssymb}
\usepackage{pifont}

\usepackage{titlesec}

\usepackage{algorithm}
\usepackage{algorithmic}
\usepackage{bm}

\newcommand\algorithmicprocedure{\textbf{procedure}}
\newcommand{\expnumber}[2]{{#1}\mathrm{e}{#2}}

\makeatletter
\newcommand\PROCEDURE[3][default]{%
  \ALC@it
  \algorithmicprocedure\ \textsc{#2}(#3)%
  \begin{ALC@prc}%
}
\newcommand\ENDPROCEDURE{%
  \end{ALC@prc}%
}
\newcommand{\floor}[1]{\left\lfloor #1 \right\rfloor}

\newenvironment{ALC@prc}{\begin{ALC@g}}{\end{ALC@g}}
\makeatother

\usepackage[T1]{fontenc}

\usepackage[utf8]{inputenc}

\usepackage{microtype}

\newcommand{\s}{\emph{shortening}}
\newcommand{\shf}{\emph{shorten factor}}

\title{Hierarchical Transformers Are More Efficient Language Models}

\author{Piotr Nawrot$^{*1}$, Szymon Tworkowski$^{*1}$, Michał Tyrolski$^1$, Łukasz Kaiser$^2$, \\
	{\bf Yuhuai Wu$^3$, Christian Szegedy$^3$, Henryk Michalewski$^3$ }\\
	$^1$University of Warsaw, $^2$OpenAI, $^3$Google Research \\
	{\small \texttt{\{p.nawrot99, szy.tworkowski, michal.tyrolski, lukaszkaiser\}@gmail.com, }} 
	\\
	{\small \texttt{\{yuhuai, szegedy, henrykm\}@google.com }}
}

\begin{document}
\maketitle

\renewcommand{\thefootnote}{\fnsymbol{footnote}}
\footnotetext[1]{Equal contribution. Order determined by coin toss.}
\renewcommand{\thefootnote}{\arabic{footnote}}

\begin{abstract}
 Transformer models yield impressive results on many NLP and sequence modeling tasks. Remarkably, Transformers can handle long sequences, which allows them to produce long coherent outputs: entire paragraphs produced by GPT-3 or well-structured images produced by DALL-E. These large language models are impressive but also very inefficient and costly, which limits their applications and accessibility. We postulate that having an explicit hierarchical architecture is the key to Transformers that efficiently handle long sequences. To verify this claim, we first study different ways to downsample and upsample activations in Transformers so as to make them hierarchical. We use the best performing upsampling and downsampling layers to create Hourglass - a hierarchical Transformer language model. Hourglass improves upon the Transformer baseline given the same amount of computation and can yield the same results as Transformers more efficiently. In particular, Hourglass sets new state-of-the-art for Transformer models on the ImageNet32 generation task and improves language modeling efficiency on the widely studied enwik8 benchmark.
\end{abstract}

\section{Introduction}

Transformer models \cite{vaswani2017attention} are capable of
solving many sequence modeling tasks, including classical NLP
tasks \cite{devlin2019bert}, summarization \cite{zhang2020pegasus}, language modeling \cite{radford2019language, brown2020language}, code generation \cite{chen2021evaluating}, or even music generation \cite{huang2018music, dhariwal2020jukebox} and image generation \cite{parmar2018image, pmlr-v119-chen20s, ramesh2021zeroshot}.
One compelling feature of Transformers is their ability
to handle long contexts given as part of the input. This is particularly visible in tasks where the output depends on parts of the context that may not be close-by in the generated sequence, like in
summarization, where the summary may need to refer to information scattered across the context, or in large-scale image generation, where pixels belonging to the same object may be far apart in the generation order. Transformers excel at such tasks thanks to self-attention, and they are used with longer and longer contexts. 

\begin{figure}[ht!]
\centering
    \resizebox{1.0\linewidth}{!}{\includegraphics{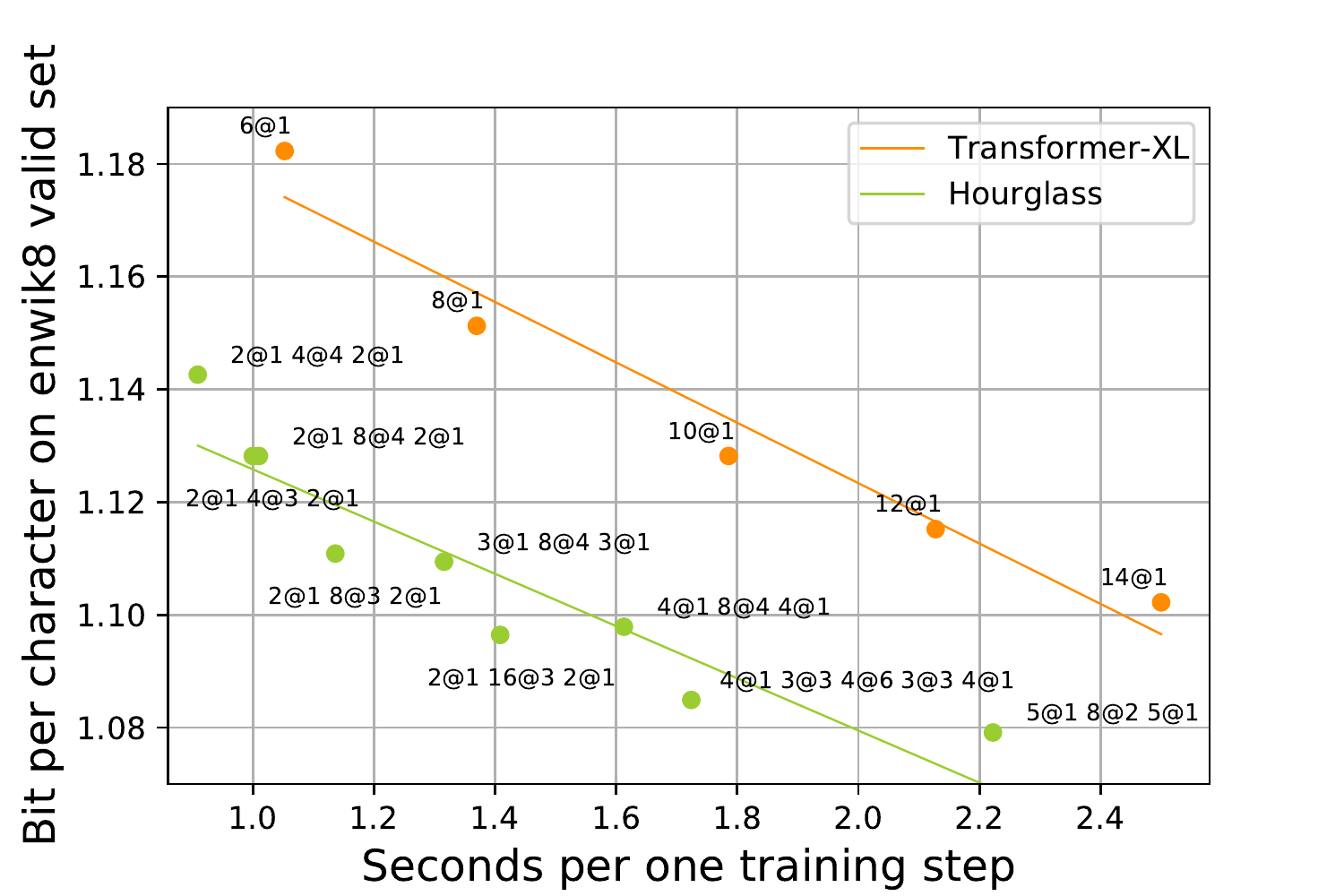}}
\centering
\caption{Bits-per-character vs.
training cost for baseline (orange) and hierarchical Transformers (green).
We observe significant perplexity improvements on enwik8 over the vanilla Transformer–XL baseline, see text for details.
}
\label{fig:linear}
\end{figure}

The ability of Transformers to handle long contexts comes at a price:
each self-attention layer, at least in its original form, has complexity quadratic in the length of the context. When a stack of $n$ Transformer layers is used, both memory and time complexity is equal to $O(L^2 n)$ where $L$ is a sequence length and $n$ number of decoder blocks. Due to this limitation, vanilla transformers are infeasible to train on tasks with very long input sequences, for instance, on high-resolution images.
This issue has been studied extensively, and a number of techniques were introduced that modify attention mechanism without changing overall transformer architecture \cite{child2019generating,roy2020efficient,ren2021combiner}. These sparse attention mechanisms reduce the complexity of self-attention but still force the model to operate on the sequence of the same length as the input.

\begin{figure*}[t]
  \centering
  \includegraphics[width=1.0\linewidth]{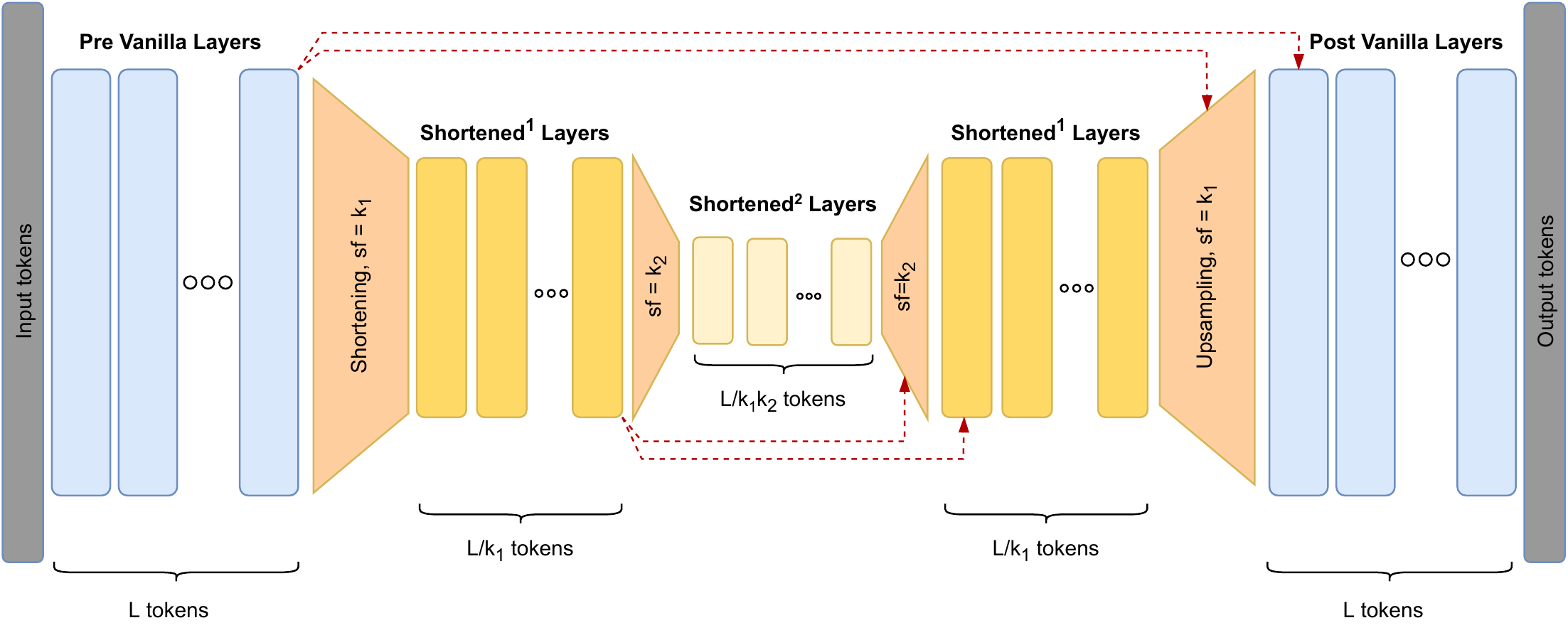}
  \caption{Hourglass - a high-level architecture overview. The arrows denote residual connections.}
  \label{fig:model}
\end{figure*}

For generative Transformer models, operating at the original scale of the input sequence is necessary, at least in the early and final layers, as the input must be processed at first and generated at the end (Section \ref{sec:vanilla}). But forcing the models to operate at this granularity throughout the layer stack has both fundamental and practical shortcomings:
\begin{itemize}
    \item Fundamentally, we aim for the models to create high-level
      representations of words, entities, or even whole events -- which
      occur at a very different granularity than single letters that
      the model receives on input.
    \item On the practical side, even layers with linear complexity
      can be slow and memory-intensive when processing very long sequences.
\end{itemize}
To alleviate these issues, we propose to change the Transformer architecture to first shorten the internal sequence of activations when going deeper in the layer stack and then expand it back before generation. We merge tokens into groups using a shortening operation (Section \ref{sec:shortening}) and so reduce the overall sequence length, and then up-sample them again combining with the sequence from earlier layers (Section \ref{sec:upsampling-methods}), The first part is analogous to the Funnel-Transformer architecture~\cite{dai2020funneltransformer}, and the whole architecture takes inspiration from U-Nets \cite{ronneberger2015unet}. In contrast to both these architectures, the model we present is autoregressive, which is harder to ensure in hierarchical models than in vanilla Transformers.

The resulting model -- which we call \emph{Hourglass} -- is an autoregressive Transformer language model that operates on shortened sequences. It yields significant performance improvements for different attention types (Fig. \ref{fig:enwik},\ref{fig:reformer}). We tested Hourglass with Transformer-XL \cite{dai2019transformerxl} and Reformer \cite{kitaev2020reformer} blocks on enwik8 dataset. In both cases, it is not only better in terms of perplexity, but it is faster and uses less memory during training. We also propose a regularization technique for hierarchical Transformers called \emph{shorten factor dropout} which improves perplexity upon baselines trained with fixed shorten factor (see Section \ref{sec:sfdropout}). Finally, Hourglass achieves the new state-of-the-art among Transformer models for image generation of ImageNet32 (see Tab.~\ref{tab:imagesota}).

\section{Model}

Standard self-attention mechanism uses full token-level sequence representations. In the Hourglass, we bring efficiency to the model by utilizing \s, which allows us to use the Transformer layers on inputs with significantly smaller lengths. A high-level overview of our proposed model architecture is shown in figures  \ref{fig:model} and \ref{fig:alg}. 

\begin{figure}[ht] 
\begin{algorithm}[H]
\begin{algorithmic}
\PROCEDURE[]{Hourglass}{$x, [k, ...s\_factors]$}
\STATE $x \gets PreVanillaLayers(x)$
\STATE $x' \gets Shortening(ShiftRight(x, k-1), k)$
\IF {\textsc{empty}($s\_factors$)}
    \STATE $x' \gets ShortenedLayers(x')$
\ELSE
    \STATE $x' \gets \textsc{Hourglass}(x', s\_factors)$
\ENDIF
\STATE $x \gets x + Upsampling(x, x', k)$
\STATE $x \gets PostVanillaLayers(x) $
\RETURN $x$
\ENDPROCEDURE
\end{algorithmic}
\caption{HourglassLM}
\end{algorithm}
\caption{
The architecture starts with \emph{pre vanilla layers} -- a stack of Transformer blocks operating on the full token-level sequence. After them we insert \emph{shortening layer} where $k$ is the \emph{shorten factor} parameter (Fig. \ref{fig:shortening}). The sequence is shifted right before shortening to prevent information leak (Fig. \ref{fig:leak}). Then we recursively insert another Hourglass block operating on $k$ times smaller scale. On the final level of shortening, we apply \emph{shortened layers} -- Transformer blocks operating on the smallest scale.
\emph{Upsampling layer} brings the resulting activations $x'$ back to the original resolution. After upsampling and residual, the activations are processed by token-level \emph{post vanilla layers}.
}
\label{fig:alg}
\end{figure}

Attention type in the vanilla layers and shortened layers is a configurable parameter. By default we use relative attention defined in Transformer-XL \cite{dai2019transformerxl}. Any attention module can be used - we show significant efficiency gains when applying Hourglass also for LSH~\cite{kitaev2020reformer} attention (see Section \ref{sec:hourglassapplication} and Fig. \ref{fig:reformer}).

\subsection{Methods of shortening the input sequence}\label{sec:shortening}

Shortening can be defined as any function $S$ that accepts a tensor $x$ of shape $(l, d)$ and returns a tensor $x'$ of shape $(\frac{l}{k}, d)$, where $k$ is a hyperparameter called \shf. 
    
A simple shortening method is 1D average pooling with stride $k$ and pool size $k$, applied along the sequence dimension $l$.
Another way of shortening is what we will further call \emph{linear pooling} ($l$ and $d$ denote sequence length and $d_{model}$):
\begin{algorithm}[H]
\begin{algorithmic}
\STATE $x' \gets Reshape(x, (\frac{l}{k}, k \cdot d))$
\STATE $x' \gets LinearProjection(x')$
\end{algorithmic}
\caption{LinearPooling}
\end{algorithm}

Shortening can be also performed by attention, as was introduced in \cite{dai2020funneltransformer}: $x' = S(x) + Attention(Q = S(x), K = V = x)$ where $S$ is shortening function, originally $S = AvgPool$. Directly after this attention operation, a positionwise feed-forward with a residual is performed, so that these two layers form a Transformer block \cite{vaswani2017attention}. In this work we also try $S = LinearPool$ and find it more effective on image tasks (see Tab. \ref{tab:pooling}).

\subsection{Shortening and autoregressive property}\label{sec:lm}

\begin{figure}
\centering
    \resizebox{0.95\linewidth}{!}{\includegraphics{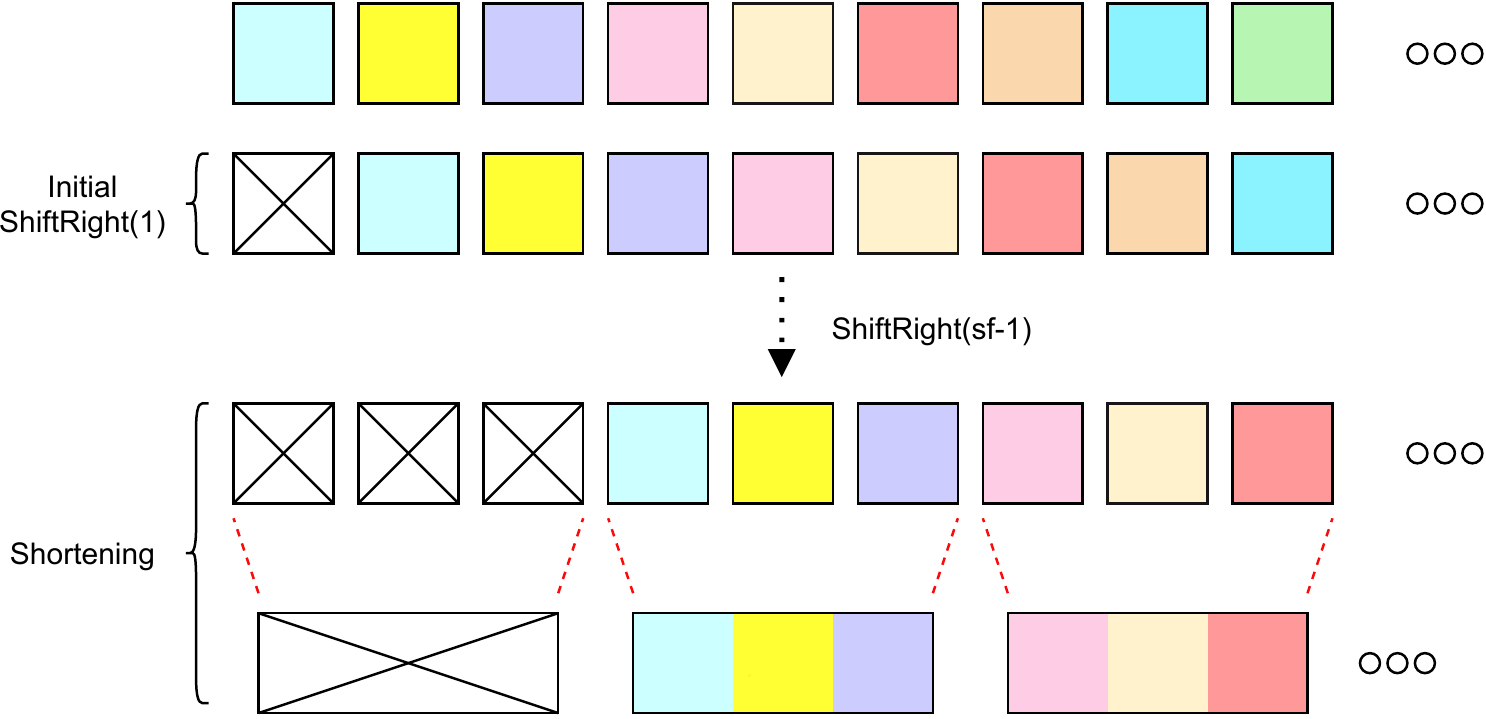}}
\centering
\caption{An overview of our shortening approach. Different colors denote token positions. Initially, we shift right by one, which is a standard step in TransformerLM. Then, just before performing shortening, we additionally shift the tokens right by $\emph{shorten factor}-1$ to preserve the autoregressive property of the model.}
\label{fig:shortening}
\end{figure}

\paragraph{Information leaks}
Shortening interferes with the standard causal masking used in Transformer decoders. 
Namely, in any shortened representation by a factor of $k$
each shortened token contributes to predicting up to the next $k$ tokens in the finest scale, 
that is if $e$ is the shortened sequence and $x$ is the sequence on the finest scale, 
$e_0$ is not only used to generate $x_0$; in fact, the same embedding is used to generate tokens $x_0, ..., x_{k-1}$.

\begin{figure}[ht]
\centering
    \resizebox{0.95\linewidth}{!}{\includegraphics{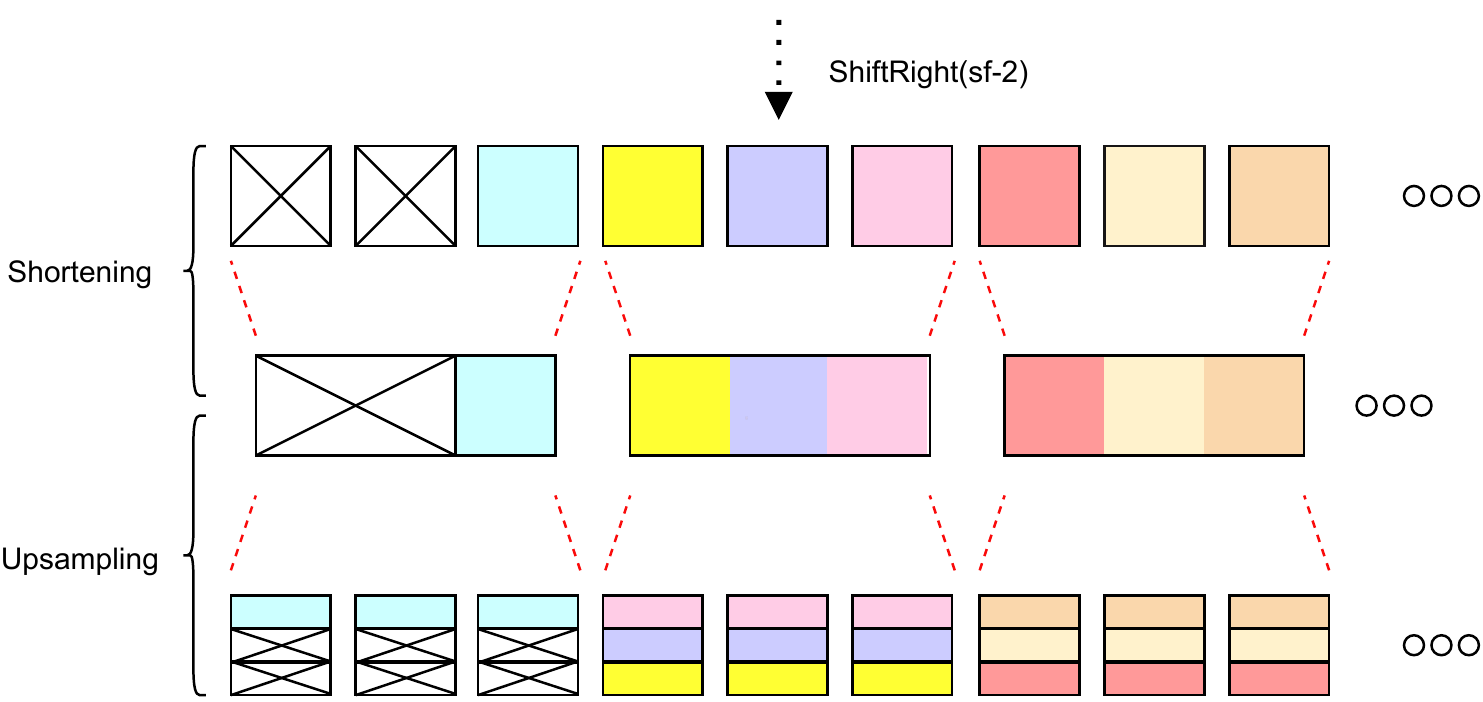}}
\centering
\caption{An example of information leak. If the shift right factor is too small, after upsampling the knowledge from the next tokens leaks to previous ones violating autoregressiveness and making decoding impossible. }
\label{fig:leak}
\end{figure}

Therefore, we need to guarantee that $e_0$ and any other $e_i$ cannot access information about tokens they will implicitly predict. To ensure that, we apply another shift right by $k - 1$ tokens, directly before any shortening by a factor of $k$ (Fig. \ref{fig:shortening}). The shift is the smallest that does not cause an information leak (see Fig. \ref{fig:leak} for an example of a shifting that leads to a leak). We included a more detailed analysis of this fact in the Appendix (Section \ref{sec:appendix.leak}).

\paragraph{Reduced expressivity}
Let us consider an Hourglass model with shortening by a factor of $k$ and no transformer blocks operating on the finest scale (that is, a model without vanilla layers). 

In this situation\\ 
$P(x) = \prod_{i=0}^{n-1}P(x_i | e_0,...,e_{\floor{\frac{i}{k}}}) = \prod_{i=0}^{n-1}P(x_i | x_0,...,x_{\floor{\frac{i}{k}} \cdot k - 1})$

because for predicting $x_i$ we combine the processing done on shortened representations $e$ with token-independent operations. This  means token $x_i$ is generated independently from the tokens $x_{\floor{\frac{i}{k}} \cdot k}, ..., x_{i-1}$. This situation is detrimental to the model's capabilities, though including at least one vanilla layer solves this issue. In the Appendix we provide a detailed example illustrating this problem (Section \ref{sec:appendix.reduced}). 

\subsection{Upsampling methods}\label{sec:upsampling-methods}
Upsampling is a crucial part of the Hourglass architecture since we need to convert shortened representations back to the full token-level sequence in order to perform language modeling.

A method proposed in \cite{dai2020funneltransformer} is repeating each shortened vector \emph{shorten factor} times. This method is computationally efficient, but it does not distinguish tokens with respect to position inside the group. 

Another method is \emph{linear upsampling} which works analogously to linear pooling -- it projects vectors of shape $(\frac{l}{k}, d)$ to $(\frac{l}{k}, k \cdot d)$ and then reshapes to $l$ vectors, each of dimension~$d$. This method is fast and allows to project shortened embeddings differently for each position in the group. This happens because the $(k \cdot d) \times d$ projection matrix can be thought of as $k$ separate $d \times d$ matrices, one per each position. 

We also investigated a method which we further call \emph{attention upsampling}. It is similar to attention pooling \cite{dai2020funneltransformer} and to the aggregation layer from \cite{subramanian2020multiscale}. It works as follows: $x = U(x, x') + Attention(Q = U(x, x'), K = V = x')$ where $x$ are embeddings from just before the shortening, $x'$ are final shortened embeddings and $U$ is an arbitrary upsampling function. After the attention operation there is also a residual with a feed-forward layer.

Linear upsampling learns a fixed pattern that is the same for each shortened token. Attention upsampling has the advantage of being content-based -- each token can extract relevant information from the shortened embeddings. We set $U(x, x') = x + LinearUpsampling(x')$ which allows to explicitly inject group-level information into the attention queries. We experimentally show that variants of attention upsampling lead to the best results for our model across different datasets (see Tab. \ref{tab:upsampling}).

\section{Experiments}

In this section, we present experimental results of Hourglass. We start with a quick analysis of time and memory complexity of the approach (Section \ref{sec:cost}). Then we investigate the efficiency gains of applying Hourglass to Transformers with different attention types (Section \ref{sec:hourglassapplication}). Finally, we use Hourglass with relative attention parametrization from Transformer-XL~\cite{dai2019transformerxl}, evaluate it on three language modeling tasks, and compare the results with other models. (Sections \ref{sec:enwik}, \ref{sec:im})

To show cross-domain generalization of our method, we train our model on one dataset related to Natural Language Processing and two from the Computer Vision field.

To ensure consistency in presenting configurations of our model, we introduce a notation describing hierarchy of our architecture: $(N_1@f_1, \dots, N_k@f_k)$ where each entry $(N_j@f_j)$ means $N_j$ layers shortened by factor $f_j$. 

\renewcommand{\footnotesize}{\fontsize{5pt}{11pt}\selectfont} 
Our model implementation is open source.\footnote{\texttt{github.com/google/trax/blob/master/trax/models/research/hourglass.py}}

\subsection{Computational cost analysis}\label{sec:cost}
In vanilla Transformers, the number of parameters can indicate the computation required to train the model. This is not true for Hourglass -- for instance, it can have 128 layers operating on a sequence shortened by 32 and still fit into the memory of a single GPU. A weak correlation between true Hourglass' computational cost and its number of parameters can be observed in Table \ref{tab:memspeed}.

Hourglass achieves the biggest speedup with the standard $\mathcal{O}(l^2)$ attention. In that case, a single shortening by a shorten factor $k$ reduces the complexity to $\mathcal{O}(\frac{l^2}{k^2})$ so by a factor of $k^2$.
For more recent linear-time attention mechanisms \cite{katharopoulos2020transformers, choromanski2021rethinking}
the reduction would be smaller -- but still by a factor of $k$.
Feed-forward layers also have linear complexity so shortening
reduces it by a factor of $k$. 

In Table \ref{tab:memspeed} we show an empirical efficiency comparison between
Hourglass and Transformer-XL.

\begin{table}[ht!]
\small
\centering
\setlength{\tabcolsep}{0.35em}
\begin{tabular}{lcccc}
\hline
Hierarchy & BPC & GB & Speed & \#Param \\
\hline
6@1 (Baseline) & $1.182$ & $4.53$ & $0.95$ & 21M\\
2@1 1@3 2@1 & $1.163$ & $4.41$ & $1.11$ & 24M\\
2@1 4@4 2@1 & $\textbf{1.143}$ & $4.41$ & $1.10$ & 34M\\
\hline
8@1 (Baseline) & $1.151$ & $5.75$ & $0.73$ & 28M\\
2@1 4@3 2@1 & $1.128$ & $4.88$ & $1.00$ & 34M\\
2@1 8@4 2@1 & $1.128$ & $4.98$ & $0.99$ & 48M\\
2@1 1@2 4@4 1@2 2@1 & $1.115$ & $4.69$ & $0.86$ & 48M\\ 
2@1 8@3 2@1 & $\textbf{1.111}$ & $5.50$ & $0.88$ & 48M\\
\hline
10@1 (Baseline) & $1.128$ & $6.99$ & $0.56$ & 34M\\
3@1 8@4 3@1 & $\textbf{1.109}$ & $6.14$ & $0.76$ & 55M\\
\hline
12@1 (Baseline) & $1.115$ & $8.12$ & $0.47$ & 41M\\
4@1 8@4 4@1 & $1.098$ & $7.20$ & $0.62$ & 62M\\
2@1 16@3 2@1 & $\textbf{1.096}$ & $5.89$ & $0.71$ & 75M\\
\hline
14@1 (Baseline) & $1.102$ & $9.35$ & $0.40$ & 48M\\
5@1 8@2 5@1 & $\textbf{1.079}$ & $9.57$ & $0.45$  & 69M\\
\hline
\end{tabular}
\caption{Efficiency comparison between Hourglass variants and Transformer-XL baseline on enwik8 -- we report validation set perplexity (BPC), running memory (GB) and number of training steps per second (Speed). We observe significant perplexity gains over the baseline for a matching computation cost. It is also visible that for Hourglass the number of model parameters (\#Param) correlates poorly with true computational cost. }
\label{tab:memspeed}
\end{table}

\subsection{Impact of Hourglass}\label{sec:hourglassapplication}
\begin{figure}[ht!]
\centering
    \resizebox{1.0\linewidth}{!}{\includegraphics{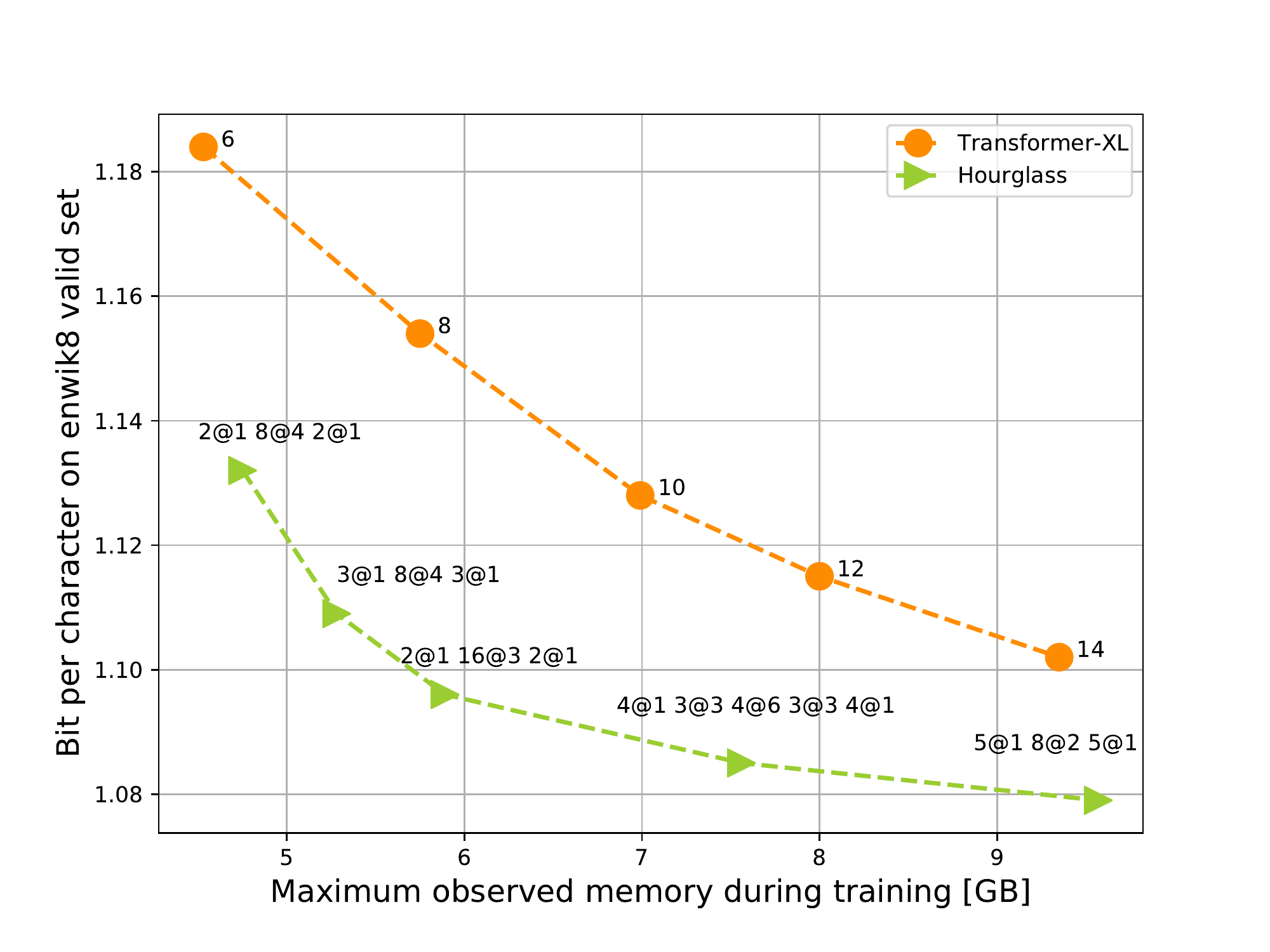}}
\centering
\caption{
Comparison between Transformer-XL baseline and Hourglass on Enwik8 valid set w.r.t. maximum memory used during training. All models are trained for 200k steps with the same hyperparameters. 
}
\label{fig:enwik}
\end{figure}

\begin{figure}
\centering
    \resizebox{1.0\linewidth}{!}{\includegraphics{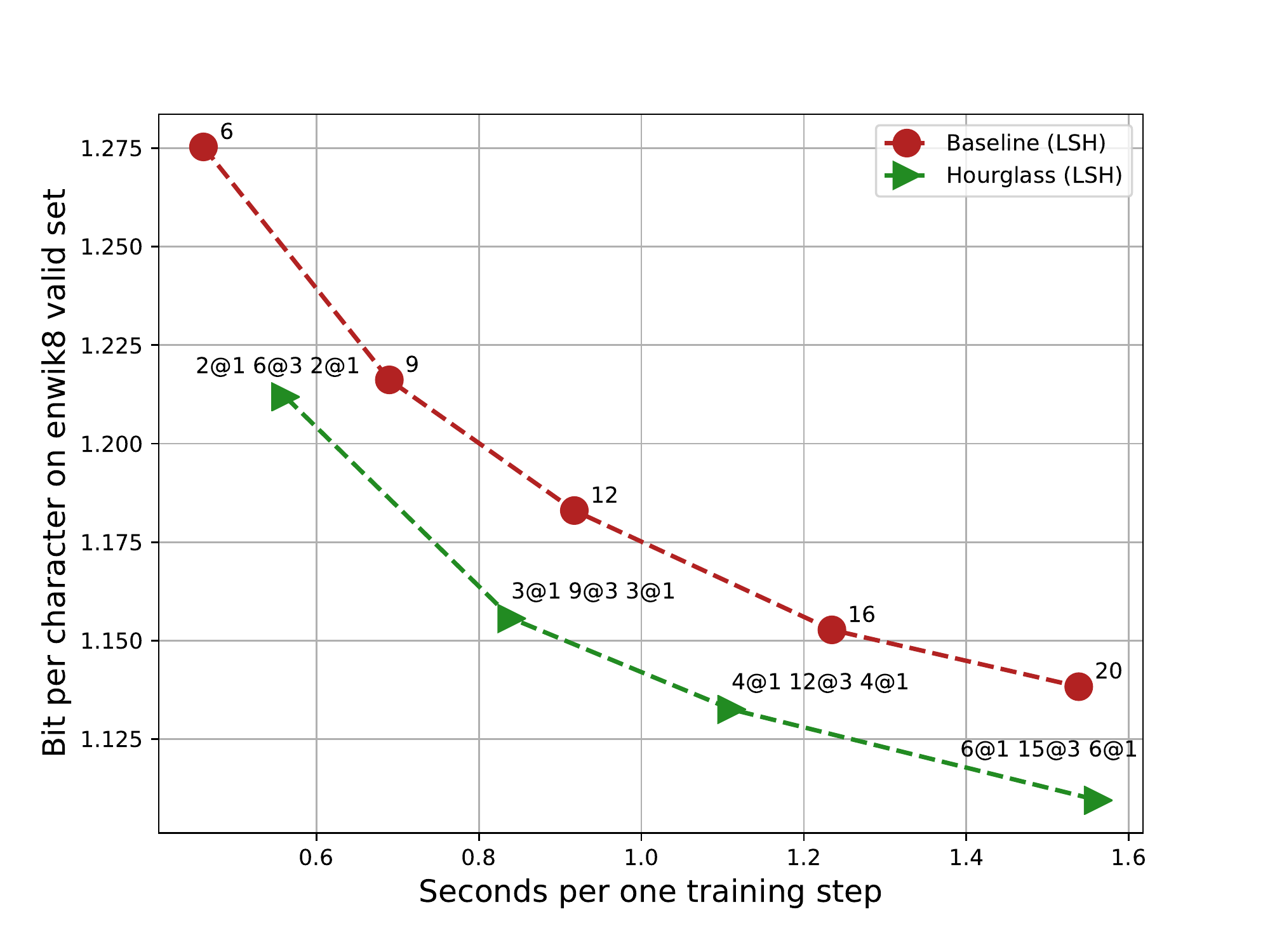}}
\centering
\caption{
Comparison between Reformer baseline and Hourglass, both with LSH attention, on Enwik8 valid set w.r.t. cost of one training step in seconds.
}
\label{fig:reformer}
\end{figure}

To demonstrate the efficiency of Hourglass, we measured how computational cost decreases and perplexity improves, purely adding the technique to Transformer-XL~\cite{dai2019transformerxl} and Reformer~\cite{kitaev2020reformer} backbones (results depicted in Figures \ref{fig:enwik} and \ref{fig:reformer}, respectively).

In both cases, models are implemented under the same codebase and the only difference between Hourglass and its corresponding baseline is the usage of shortening and upsampling layers. We show that by incorporating a single shortening of the input, we can train larger models with the same memory requirements and training speed and achieve better perplexity than baselines.

\setlength{\textfloatsep}{0.5em}
\subsection{Enwik8}\label{sec:enwik}
Enwik8 \cite{enwik} is a byte-level language modeling benchmark containing the first 100M bytes of unprocessed English Wikipedia text, split into 90M train, 5M valid, and 5M test sets.

Similarly to \cite{dai2019transformerxl} and \cite{beltagy2020longformer}, we evaluate our model on the test set, splitting it into overlapping sequences of size $l = 4096$ with a step size of $128$ and calculate the test loss only over the last $128$ tokens. With a $(4@1, 8@3, 4@1)$ hierarchy, $d_{model}=768$, $d_{ff}=3072$ and $8$ heads, we reach \textbf{0.98} test bits-per-character.

\begin{table}[t]
\setlength{\tabcolsep}{1em}
\small
\centering
    \begin{tabular}{lcc}
    \hline
    \textbf{Enwik8} & \#Param & BPC \\
    \hline
    Transformer-XL
    \shortcite{dai2019transformerxl} 24L~ & 277M & 0.99 \\
    \textbf{Hourglass} & 146M & 0.98 \\
    \hline
    Adaptive-Span \shortcite{sukhbaatar2019adaptive} 24L~ & 209M & 0.98 \\
    Transformer-LS \shortcite{zhu2021longshort} & 110M & 0.97 \\ 
    Feedback Transformer \shortcite{fan2021addressing} & 77M & 0.96 \\ 
    Expire-Span \shortcite{sukhbaatar2021memories} 24L~ & 277M & \bf 0.95 \\ 
    \hline
    \end{tabular}
    \caption{
    \textbf{Enwik8 Results.} We report bits-per-character (BPC) on the test set and number of model parameters. Hourglass applied to Transformer-XL significantly outperforms its baseline. Our technique could be also used with other more performant attention methods which we leave for future work.
    }
    \label{tab:enwik8}
\end{table}

\subsection{Image Generation}\label{sec:im} We use datasets introduced in \cite{DBLP:journals/corr/OordKK16} which are downsampled versions of the popular ImageNet. In the autoregressive image generation setup, they consist of respectively $\bm{32\times32\times3}$ and $\bm{64\times64\times3}$ tokens, corresponding to RGB channels, per image. As the only preprocessing step we flatten the images.

\subsubsection{ImageNet32}
For our main result the following hierarchy is used: ($3@1, 24@3, 3@1$). We use $d_{\mathrm{model}} = 512$, $d_{\mathrm{ff}} = 2048$, $8$ attention heads and $0.01$ dropout rate. With this configuration we achieve $\textbf{3.741}$ bits/dim, yielding the new state-of-the-art among autoregressive (Transformer-based) models on this dataset, compared to the previous state-of-the-art of 3.758 bpd by \cite{ho2019axial}. 

\begin{figure}
\centering
    \resizebox{1\linewidth}{!}{\includegraphics{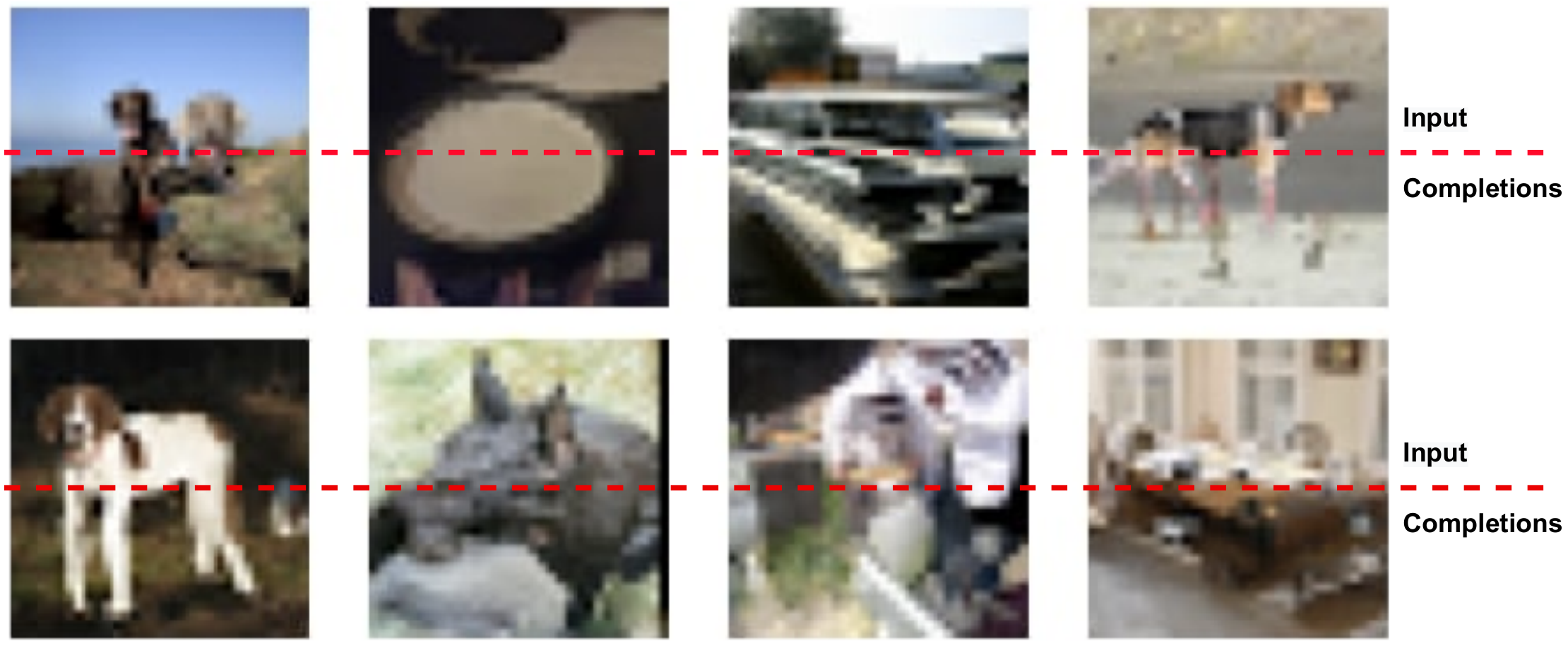}}
\centering
\caption{Examples of our model completions, where bottom half of each image was generated by our model, prompted by the upper half.}
\label{fig:im32}
\end{figure}

\begin{table} 
\small
\centering
\setlength{\tabcolsep}{1.5em}
\begin{tabular}{lc}
\hline
\textbf{ImageNet32} & BPD \\
\hline
PixelCNN \cite{oord2016conditional} & 3.83 \\
Image Transformer \cite{parmar2018image} & 3.77 \\
Axial Transformer \cite{ho2019axial} & 3.76 \\
\textbf{Hourglass} & 3.74 \\
\hline
VDM \cite{kingma2021variational} & 3.72 \\ 
DenseFlow \cite{grcic2021densely} & \textbf{3.63} \\
\hline
\hline
\textbf{ImageNet64} & BPD \\
\hline
Reformer \cite{kitaev2020reformer}
& 3.65 \\
Performer \cite{choromanski2021rethinking}
& 3.64 \\
\textbf{Hourglass} 
& 3.44 \\
Sparse Transformer \cite{child2019generating}
& 3.44 \\
Routing Transformer \cite{roy2020efficient}
& 3.43 \\
Combiner \cite{ren2021combiner} 
& 3.42 \\
\hline
VDM \shortcite{kingma2021variational} & 3.40 \\
DenseFlow \shortcite{grcic2021densely} & \textbf{3.35} \\
\hline
\end{tabular}
\caption{Bits per Dimension (BPD) on downsampled imagenet. Autoregressive models are separated by a horizontal line from non-autoregressive ones. On ImageNet32, our model yields new state-of-the-art for autoregressive models. } 
\label{tab:imagesota}
\end{table}

\subsubsection{ImageNet64} \label{sec:im64}
The sequence length that our model can handle is limited mainly by the computational complexity of used attention module. 
We replace relative attention in vanilla layers by LSH attention \cite{kitaev2020reformer}, which allows us to handle $\bm{12288}$-long sequences. To achieve relative attention parametrization, the LSH attention is combined with rotary positional embeddings \cite{su2021roformer}. In shortened layers, standard relative attention is used. For LSH attention, we set chunk length to $128$ and use $2$ hashes, which results in small memory complexity in our full-size layers. In this setup, we reach a score of $\textbf{3.443}$ bpd with a $(3@1, 12@3, 3@1)$ architecture. All attention layers had $d_{model} = 768$, $d_{ff} = 3072$ and 8 heads. No dropout was used. 
\subsubsection{CIFAR-10} 
CIFAR-10 \cite{Krizhevsky09learningmultiple} is an image dataset consisting of 60000 images of size 32x32. We use this dataset primarily for our ablations (Section \ref{sec:ablation}). Due to the relatively small number of examples compared to ImageNet, models reach convergence after 100k steps.

\setlength{\textfloatsep}{0.5em}
\section{Ablations}\label{sec:ablation}
In this section, we start by introducing a training technique called \emph{shorten factor dropout} (Section \ref{sec:sfdropout}), and then analyze Hourglass's components described above. We show that shortened layers behave similarly to full token-level layers in terms of scalability (Section \ref{sec:scaling}). Then we study the effect of different distributions of $(pre, post)$ vanilla layers on Hourglass' accuracy (Section \ref{sec:vanilla}). We further analyze the performance of various upsampling and downsampling methods (Sections \ref{sec:upsampling} and \ref{sec:pooling}). Finally, we discuss different shorten factors and multi-stage shortening in Section \ref{sec:sf}. 

We conduct the ablations on both text and image generation to show applicability across different domains.
We report bits per character (BPC) on the enwik8 validation (dev) set evaluated without context (sequence length $2048$) and bits per dim (BPD) on the CIFAR-10 test set.
For the exact hyperparameter setup refer to the Appendix.

\subsection{Shorten factor dropout}\label{sec:sfdropout}

Different shorten factors can be used for the same model when using parameterless pooling methods. We propose a training procedure where the shorten factor is randomly sampled with uniform distribution from a predefined set in each step. We observe that such a training regime improves validation loss compared to a baseline trained with a single, fixed shorten factor.
For example, a model trained with shorten factor randomly sampled from $\{2,3\}$ performs better when evaluated with any of these shorten factors, compared to models trained with a corresponding fixed shorten factor (Tab. \ref{tab:sfdropout}).

We hypothesise that such a technique promotes a more uniform distribution of information over the sequence of tokens. It may be essential for fixed-size pooling techniques as they do not account for variable length constituents like words. By spreading information uniformly, we prevent a situation where we lose content by shortening three information-dense tokens or lose available capacity by merging three low information ones.

Shorten factor dropout is not limited to our architecture and can be applied to any model that utilizes shortening, particularly \cite{dai2020funneltransformer}.

\begin{table}[ht!]
\small
\centering
\setlength{\tabcolsep}{0.8em}
\begin{tabular}{lrrr}
\hline
Hierarchy & Train $k$ & Val $k=2$ & Val $k=3$ \\
\hline
2@1 8@$k$ 2@1 & $\{2,3\}$ & $\textbf{1.104}$ & $\textbf{1.116}$ \\
              & 2 & $1.116$ &         \\
              & 3 &         & $1.124$ \\
\hline
4@1 12@$k$ 4@1 & $\{2,3\}$ & $\textbf{1.086}$ & $\textbf{1.094}$ \\
              & 2 & $1.098$ &         \\
              & 3 &         & $1.101$ \\
\hline
5@1 10@$k$ 5@1 & $\{2,3\}$ & $\textbf{1.082}$ & $\textbf{1.087}$ \\
              & 2 & $1.096$ &         \\
              & 3 &         & $1.095$ \\
\hline
\end{tabular}
\caption{Comparison between models trained with shorten factor dropout (Train $k=\{2,3\}$, Section \ref{sec:sfdropout}) and fixed shorten factor baselines on enwik8.}
\label{tab:sfdropout}
\end{table}

\vspace{-5mm}
\subsection{Scaling shortened layers}\label{sec:scaling}
In this study, we show that layers operating on the shortened sequence contribute significantly to Hourglass's accuracy. In Table \ref{tab:scaling} we measure the impact of scaling the depth of the shortened part of the model with a fixed number of vanilla layers.

We also check if scaling laws of Transformers, described in \cite{kaplan2020scaling}, hold by comparing a regression line fitted to various Hourglass configurations and one fitted to Transformer-XL baseline. We observe in Figure~\ref{fig:linear} that the slopes are very similar, which indicates that the laws hold.

\begin{table}[ht!]
\small
\centering
\setlength{\tabcolsep}{0.95em}
\begin{tabular}{lrr}
\hline 
Number of shortened layers & enwik8 & CIFAR-10\\
\hline
Baseline ($n=1$) & $1.164$ & $3.28$ \\
\hline
$n=4$ & $1.134$ & $3.16$ \\
$n=8$ & $1.111$ & $3.07$ \\
$n=16$ & $\textbf{1.096}$ & $\textbf{3.03}$ \\
\hline
\end{tabular}
\caption{Impact of increasing the number of shortened layers on perplexity.
Vanilla layers: $(1,1)$ for CIFAR-10 and $(2,2)$ for enwik8, shorten factor $3$ used in both.
}
\label{tab:scaling}
\end{table}

\vspace{-5mm}
\subsection{Impact of vanilla layers}\label{sec:vanilla}
We observe a significant contribution to Hourglass' performance with increasing the number of vanilla layers. One reason is that we perform more computations as in vanilla layers we process the sequence in token-level - no shortening is applied. We also see that the distribution of vanilla layers before shortening and after shortening does impact the training (see Tab. \ref{tab:vanilla}), and equal distribution leads to the best perplexity.

\begin{table}[ht!]
\small
\centering
\setlength{\tabcolsep}{1.9em}
\begin{tabular}{lrr}
\hline
Vanilla layers & enwik8 & CIFAR-10 \\
\hline
(0, 0) & $1.460$ & $3.429$ \\ 
\hline
(0, 2) & $1.176$ & $3.108$ \\
(2, 0) & $1.189$ & $3.035$ \\
(1, 1) & $\textbf{1.171}$ & $\textbf{3.012}$ \\
\hline
(2, 2) & $1.128$ & $2.966$ \\
\hline
\end{tabular}
\caption{Impact of the distribution of vanilla layers on enwik8 (BPC) and CIFAR-10 score (BPD). We see that equal distribution of layers before and after shortening leads to better results on both datasets.}
\label{tab:vanilla}
\end{table}

\subsection{Upsampling method}\label{sec:upsampling}
In Table~\ref{tab:upsampling} we investigate different possibilities of choosing the upsampling method. For attention-free methods, linear upsampling performs better on images, while repeat upsampling works well for text. Attention upsampling works well regardless of the function $U$ and has the lowest perplexity.

\begin{table}[ht!]
\small
\centering
\setlength{\tabcolsep}{0.9em}
\begin{tabular}{lrr}
\hline
Upsampling method & enwik8 & CIFAR-10 \\
\hline
Repeat & $1.148$ & $3.062$ \\ 
\hline
Linear & $1.163$ & $3.020$  \\
$U(x, x') = x$ & $1.145$ & $\textbf{2.967}$ \\
$U(x, x') = x + Linear(x')$ &  $\textbf{1.132}$ & $3.012$ \\
\hline
\end{tabular}
\caption{Upsampling method ablation - baseline configurations are $(2@1, 24@4, 2@1)$ and $(1@1, 8@3, 1@1)$ for enwik8 and CIFAR-10, respectively. }
\label{tab:upsampling}
\end{table}

\vspace{-5mm}
\subsection{Pooling method}\label{sec:pooling}
Table \ref{tab:pooling} presents impact of pooling method on both enwik8 (BPC) and CIFAR-10 (BPD). Attention pooling reaches the lowest perplexity for both datasets. Average pooling performs well on text among attention-free methods, while linear pooling works better for images. Both of these methods perform significantly worse for the other modality. Attention pooling demonstrates small differences with respect to chosen shortening function $S$ (Section \ref{sec:shortening}), still preserving the preference towards linear pooling on images and average pooling on text.

\begin{table}[ht!]
\small
\centering
\setlength{\tabcolsep}{0.9em}
\begin{tabular}{lrr}
\hline
Pooling method & enwik8 & CIFAR-10\\
\hline
AvgPool & $1.129$ & $3.116$\\
\hline
Attention, $S = AvgPool$ & $\textbf{1.124}$ & 3.012 \\
Attention, $S = LinearPool$  & $1.142$ & $\textbf{2.998}$ \\
LinearPool & $1.159$ & $\textbf{2.998}$ \\
\hline
\end{tabular}
\caption{Ablation of pooling methods. Attention pooling achieves the best perplexity on both datasets.}
\label{tab:pooling}
\end{table}

\vspace{-6mm}
\subsection{Shortening strategies} \label{sec:sf}

While the analysis above gives a clear indication of what methods to choose for shortening and upsampling, we are still left with the question of which shorten factors to use and whether to do single-stage or multi-stage shortening. 

Consistently, it is beneficial to do at least one shortening and by a factor of at least 3, while keeping 2-3 vanilla layers. Beyond that, a number of different configurations can yield similar results. In Table~\ref{tab:memspeed} we present the different hierarchical configurations that we tested on enwik8 and plotted in Figure~\ref{fig:linear}. It can be seen that configurations with similar computation costs perform similarly. The sequence length used in these experiments is 2048 -- we hypothesise that more hierarchy may be beneficial with even longer sequences.

\bgroup
\setlength{\parskip}{0.0em}
\section{Related Work}
\paragraph{Shortening in Transformers} Shortening in our work is inspired by Funnel-Transformer \cite{dai2020funneltransformer}. The key difference is that they train an encoder model for text classification, where our work is entirely focused on language modeling, which provides additional challenges we had to solve regarding shortening in the autoregressive setup (Section \ref{sec:lm}). Another difference is that they use repeat upsampling method while we use attention. There are also a few works related to character-level modeling which use shortening, namely \cite{clark2021canine} and \cite{tay2021charformer}. However, the authors of these works focused mainly on shortening sequence in encoder part of the transformer, whereas we focused on applying shortening in decoder.

The idea of shortening is also discussed in \cite{subramanian2020multiscale}. However, proposed architectures either focus on downsampling or upsampling, while Hourglass is a U-Net-like architecture and is symmetric in these terms. Their models use transformer layers on the finest scales when postprocessing final representations. We do these also, in the beginning, to preprocess tokens on the finest scale, and we have found it essential to the score (Section \ref{sec:vanilla}). Our attention upsampling method is similar to their \emph{aggregation layer} in the bottom-up model, however we use one upsampling for each scale change while they combine different scales to create one global upsampling.

\paragraph{Relative positional encoding} Our work is primarily built on the backbone of Transformer-XL \cite{dai2019transformerxl} - we use the same relative attention parametrization. Instead of the segment-level recurrence mechanism, we use shortening to make our model more efficient and feasible to train on longer sequences. Another relative attention parametrization is RoFormer \cite{su2021roformer} where rotary positional embeddings are introduced. We find this work particularly relevant because the method is compatible with any attention type, including efficient attention, and can be combined with our model (Section \ref{sec:im64}).

\paragraph{Sparse Attention} A well-known approach addressing the memory bottleneck is utilizing sparsity patterns in the attention matrix - Routing \cite{roy2020efficient} and Sparse Transformer \cite{child2019generating} are examples of such methods. Our solution is different in the sense that it uses full attention - just with shortened sequence length. Combiner \cite{ren2021combiner} makes a step further and provides full attention capabilities with similar computational complexity to Routing and Sparse transformers by leveraging structured factorization. This work, similarly to papers mentioned above on efficient transformers, concentrates on speeding up the attention component, while the most important feature of the Hourglass architecture is that it can use any attention module as a drop-in.

\paragraph{Image generation on downsampled ImageNet} VDM~\cite{kingma2021variational} and DenseFlow~\cite{grcic2021densely}
are recently proposed state-of-the-art methods for density estimation on this dataset. The difference between these methods and Transformer-based methods \cite{parmar2018image, ho2019axial} including this work is that the former, unlike Transformers, are non-autoregressive. 

\section{Conclusion}
In this paper, we show how hierarchy can improve the efficiency of Transformers in a language modeling setup. Our proposed architecture, Hourglass, significantly outperforms the baseline both in terms of perplexity reached at a given computation cost 
(Figure~\ref{fig:linear}) and empirical metrics like running memory (Figure~\ref{fig:enwik}). Hourglass achieves state-of-the-art results among autoregressive models on the  ImageNet32 generation task and competitive results on other image generation and language modeling tasks.

Hourglass can be used with any attention type, which opens many directions for future research related to Transformers capable of processing longer sequences. Another line of future work might be related to advances in the shortening mechanism itself, for example, involving a dynamic pooling operation that could explicitly handle the problem of fixed-size groups in multi-stage shortening.
We also leave open the problem of choosing the best hierarchy for a task. We conjecture that experiments with much longer contexts will provide better guidance for this choice and will benefit even more from the Hourglass architecture. 

\section{Acknowledgments}
Some experiments were performed using the Entropy cluster funded by NVIDIA, Intel, the Polish National Science Center grant UMO-2017/26/E/ST6/00622, and  ERC Starting Grant TOTAL. The work of Henryk Michalewski was supported by the Polish National Science Center grant UMO-2018/29/B/ST6/02959. The authors would like to thank Marek Cygan and Kamil Wilczek for their help with cluster setup, and Grzegorz Grudziński, Dawid Jamka and Sebastian Jaszczur for helpful discussions. This article describes a Team Programming Project completed at the University of Warsaw in the academic year 20/21. We are grateful to Janusz Jabłonowski, the head of Team Programming Projects,  for his support and open-mindedness.

\egroup 

\bibliography{anthology,custom}
\bibliographystyle{acl_natbib}

\clearpage
\appendix
\section{Autoregressive shortening}

In Section 2.2 we address two problems of shortening in an autoregressive setup: information leaks and reduced expressivity. Here we study these issues in more detail.

\subsection{Motivation behind using vanilla layers}\label{sec:appendix.reduced}

At first sight, it may be tempting to create hierarchical models that directly shorten the input to maximize the efficiency gains. In this section, we explain why vanilla layers are crucial for modeling at least some sequences, especially due to autoregressivity.

Consider a sequence modeling task where the input is a random sequence with repeats, such as \texttt{A\#AC\#CD\#DB\#B}. The sequence consists of chunks \texttt{L\#L} where \texttt{L} is a random uniform letter and \texttt{\#} is a special symbol. A vanilla Transformer language model can achieve $66\%$  sequence accuracy on this task -- it cannot predict the token at the beginning of the chunk, but it can predict the last token of the chunk by simply copying the token at $2$ positions earlier, which is possible using a vanilla self-attention layer.

It is however not easy to learn this task in a shortening setup when there are no vanilla layers operating on the finest scale -- this is the situation defined in \emph{Reduced expressivity} subsection of Section 2.2. Assume shorten factor is $k = 3$ and the input is \texttt{A\#AB\#BC\#C}. To avoid information leak, we shift the input sequence right by $1$, and then by $k - 1 = 2$ directly before shortening. Then the sequence is \texttt{000A\#AB\#B}. Our shortened embeddings are as follows: $e_0 = S(emb_{0}, emb_{0}, emb_0), e_1 = S(emb_{A}, emb_{\#}, emb_{A})$ where $emb$ is input embedding matrix and $S$ is a shortening function. 

\begin{table}[ht!]
\small
\centering
\begin{tabular}{l|l}
\hline
Shortened embeddings & \texttt{[000][A\#A][B\#B]} \\
Shifted input embeddings & \texttt{ 0A\#\space\space AB\#\space\space BC\# } \\
\hline
Target sequence & \texttt{ A\#A\space\space B\#B\space\space C\#C} \\
\hline
\hline
Positions & \texttt{ 123\space\space 456\space\space 789} \\
\hline
\end{tabular}
\caption{Example input sequence which is difficult to model without vanilla layers. The model can use only input embeddings shifted by one from the residual and shortened embeddings (shorten factor is 3) to predict the target sequence. Note that it is impossible to predict tokens at positions divisible by 3 using only that information.}
\label{tab:toytask}
\end{table}

Because no vanilla layers are used, for prediction we can use only shortened embeddings and input token embeddings shifted right by 1 from the residual connection. Note that to predict the \texttt{A} token at position $3$ we can use only embedding of $emb_{\#}$ and $e_0$ - both of these contain no information so we are unable to predict this token better than randomly (see Table \ref{tab:toytask}). An analogous situation occurs for prediction of any tokens at positions divisible by $3$, which makes the model unable to achieve more than $50\%$ accuracy when the task has vocabulary size of at least $2$.

This issue can be solved by adding at least one vanilla layer to the model, so that it can attend within the neighborhood of $k$ previous tokens. For this particular problem, it is sufficient to use local attention with context size $k$ in vanilla layers which is significantly more efficient than full attention.

\subsection{Information leaks -- analysis}\label{sec:appendix.leak}
\subsubsection{Definition of autoregressive model}
Formally, given a target sequence, $x = x_1, ..., x_n$, an autoregressive model (e.g. transformer decoder) models the sequence as $P(x) = \prod_{i=1}^{n} P(x_i | x_1,...,x_{i-1})$ and $$\forall_i P(x_i | x_1,...,x_n) = P(x_i | x_1,...,x_{i-1})$$ namely $x_i$ token depends only on previous tokens, never on itself nor next ones.

\subsubsection{Definition of information leak}
We say that a leak was caused by function $F_n\colon A^n \longrightarrow A^n$ transforming sequence of input tokens $(x_1, x_2, ... ,x_n)$ into another sequence $(u_1, ..., u_n) = F((x_1,..., x_n))$ when $\exists_{i<j<n} P(x_i | x_1, ..., x_{i-1}, x_j) \neq P(x_i | x_1, ..., x_{i-1})$, namely there exists $j \ge i$ that token $x_i$ depends on $x_j$ which violates the autoregressive property.

\subsubsection{Model representation}

Let $R_k\colon A^n \longrightarrow A^n$ be a shift right function which reindexes tokens by shifting each of them right by $k$ positions:

$$R_k((x_1,x_2,...x_n)) = (\underbrace{0,...,0}_{k},x_1,...,x_{n-k})$$

$S_k: A^* \longrightarrow A^*$ shortening function with factor $k$ which takes on input $x_1, ..., x_n$ sequence and returns $s_1, ..., s_m$ where $m=\frac{n}{k}$, $U_k$ upsampling function which works in similar way but upsamples $U_k((u_1,...,u_m))=u_1,...,u_n$.

Between them there is also applied $D$ decoder function, $D = D_1 \circ \dots \circ D_l$, where each $D_i$ is a function representing decoder block. Due to causal attention masking in the decoder block, there is no risk of information leak caused by function $D$. 

\subsubsection{Leak description}
Because of mentioned attention mask, we will omit the flow of information between tokens caused by the influence of attention mechanism because this mask keeps the autoregressive property.

Now, let $(x_1, ..., x_n)$ be an input sequence and $(u_1, ..., u_n) = U(D(S_k(T_s((x_1, ..., x_n))))) = F$. In order to preserve autoregressive property, it is obligatory that no leak occurs.

We will show that shift by any value $0<s<k-1$ where $k$ is the shorten factor will cause a leak.\\
To start with, consider input sequence $(x_1, ..., x_n)$ and perform operation $F$. $R_s((x_1, ..., x_n)) = (\underbrace{0,...,0}_{s},x_1,...,x_{n-s}) = r$. Assuming that $n$ is divisible by $s$, we have $S_k(r) = (v_1, ..., v_{\frac{n}{k}}) = v$ where each $v_i$ consists of information obtained in $(r_{(i-1)\cdot k + 1},..., r_{ik})$. Now let see that operation $D$ preserves autoregressive property, let $d = D(t)$. Now, $U(d) = (u_1, ..., u_n)$ and each $u_i$ depends on $d_{\floor{\frac{i-1}{k}}+1}$. 

Now consider $s \le k-2$ and let $(u_1, ..., u_n) = F((x_1,...,x_n))$ will be a result of our Transformer part. Let take $u_1$ which depends on $d_1$ and $d_1$ depends on $ (r_1, ..., r_k) = (0,...,0,x_1,...,x_{k-s}) $. For that reason $d_1$ depends on $x_{1}, x_{2}, ..., x_{k-s}$,
so we have
 $$ P(x_1 | x_{k-s}) \neq P(x_1)$$
which violates the autoregressive property.

\section{Experimental setup}
\subsection{Common parameters}
Here we list common hyperparameters used for all experiments mentioned in the paper. We use Adam optimizer with $\beta_1 = 0.9$, $\beta_2 = 0.98$ and $\epsilon = \expnumber{1}{-9}$. Weight decay and gradient clipping is not used.

In terms of model details, we decided to use a Pre-Norm architecture and FastGelu activation in feed-forward layers. 

\subsection{Enwik8}
We use $d_{model} = 512$, $d_{ff} = 2048$ and 8 attention heads. Models in ablation study are trained for 200k steps with cosine learning rate schedule, setting cycle length for 200k steps and linear warmup of $4000$ steps. 

For the main result achieving \textbf{0.98} bpc with $4@1, 8@3, 4@1$ hierarchy, we set $d_{model} = 768$, $d_{ff} = 3072$ and $n_{heads} = 8$ which results in 146M parameters. It is trained for a total number of 350k steps with one cycle of cosine schedule. Linear warmup of 20k steps is used.

At the beginning of our work on this paper, we have performed grid search over following hyperparameters for enwik8:
\begin{itemize}
    \item batch size: $\{ 8, 16, 32\}$, finally chosen $8$
    \item dropout: $\{ 0.05, 0.1, 0.15, 0.20 \}$, finally chosen $0.15$ 
    \item learning rate:\\ $\{\expnumber{1}{-4}, \expnumber{2}{-4}, \expnumber{3}{-4}, \expnumber{4}{-4}, \expnumber{5}{-4} \}$,\\ finally chosen $\expnumber{4}{-4}$
\end{itemize}
All next experiments were conducted using these parameters without additional searching.

\subsection{Downsampled ImageNet - common parameters}
For ImageNet32 and ImageNet64 experiments we use inverse square root learning rate decay from \cite{vaswani2017attention}, setting warmup steps to $8000$ in both experiments. Total batch size is $64$.

\subsection{ImageNet32}
In this dataset, we operate on input sequence length of $3072$. We use $d_{\mathrm{model}} = 512$, $d_{\mathrm{ff}} = 2048$, $8$ attention heads and $0.01$ dropout rate. We perform 400k training steps with linear warmup and inverse square root decay and then we train for additional 70k steps with cosine learning rate decay, starting from the learning rate from the previous schedule at 400k and decreasing it to $0$ at 470k steps. 

\subsection{ImageNet64}
As an input we receive a sequence of $12288$ tokens representing $64\times64\times3$ images. We set $d_{\mathrm{model}} = 768$, $d_{\mathrm{ff}} = 3072$, $8$ attention heads and dropout equal to $0$. We perform 300k steps with linear warmup and inverse square root decay.

\subsection{CIFAR-10}
All the ablation studies are run for 100k training steps, unless otherwise specified. Input sequence has length $3072$ and model parameters are as follows: $d_{\mathrm{model}} = 512$, $d_{\mathrm{ff}} = 2048$, $8$ attention heads and dropout equal to $0$. Total batch size is $8$. Cosine learning rate decay with linear warmup of $5000$ steps and 100k cycle length is used.

\section{Environment setup}
\subsection{Hardware}
Experiments are conducted on several setups. 
\begin{itemize}
    \item Ablation Study and short training sessions were computed on nodes consisting of 4x \emph{Titan V} with 12GB memory each, 64GB RAM, \emph{Intel Xeon E5-2660 v4} CPU
    \item longer trainings were completed on 8x \emph{RTX 2080 Ti} with 11GB memory each, 128GB RAM and \emph{Intel Xeon E5-2660 v4} CPU.
    \item Few longest trainings were conducted on $8\times8$ TPUv3 units, each with 16GB memory.
\end{itemize}

\subsection{Software}
All experiments were performed on Linux operating system using Trax library version 1.3.9 along with all its dependencies from this particular release date. Additionally, to run shorten factor dropout experiments we modified the Transformer-XL codebase in PyTorch.

\section{Reproducibility}
To ensure the reproducibility of this work and to support open science principles, we made our code publicly available at \texttt{github.com/google/trax}. In this repository, we also provide Google Colab notebooks where the evaluation of our main \href{https://colab.research.google.com/github/google/trax/blob/master/trax/models/research/examples/hourglass_enwik8.ipynb}{Enwik8} and \href{https://colab.research.google.com/github/google/trax/blob/master/trax/models/research/examples/hourglass_downsampled_imagenet.ipynb}{ImageNet32/64} results can be reproduced.\footnote{\texttt{https://github.com/google/trax/blob/master/trax/models/research/examples/hourglass\textunderscore enwik8.ipynb}}\footnote{\texttt{https://github.com/google/trax/blob/master/trax/models/research/examples/hourglass\textunderscore downsampled\textunderscore imagenet.ipynb}} 

\subsection{Randomness}
Seeds in all experiments were chosen randomly, however each experiment contains history which allows retrieving all randomly set parameters for reproductions. 

For each ablation described in the ablation study section, we rerun the baseline $3$ times to calculate standard deviation. All other experiments are run only once due to costs and since the variance we noticed was minimal.

\subsection{Experiment representation}
Each experiment is represented by a configuration file that unambiguously determines the whole setup -- all hyperparameters and training details like specific optimizers, data preprocessing functions, or batch size per device.

\end{document}